\documentclass[11pt]{article}

\usepackage[margin=1in]{geometry}
\usepackage[T1]{fontenc}
\usepackage[utf8]{inputenc}
\usepackage{lmodern}
\usepackage{microtype}
\usepackage{amsmath,amssymb,amsfonts}
\usepackage{graphicx}
\usepackage{booktabs}
\usepackage{tabularx}
\usepackage{multirow}
\usepackage{enumitem}
\usepackage{hyperref}
\usepackage{xcolor}
\usepackage{listings}
\usepackage{caption}
\usepackage{subcaption}

\hypersetup{
  colorlinks=true,
  linkcolor=blue!60!black,
  citecolor=blue!60!black,
  urlcolor=blue!60!black,
}

\newcommand{\cmark}{$\checkmark$}
\newcommand{\xmark}{$\times$}
\newcommand{\arxivcomment}[1]{\textcolor{gray}{\small #1}}

\DeclareMathOperator{\relu}{relu}

\title{%
  \textbf{PsiLogic: Chaos-Aware Active Cancellation for Adam\\
  with a Fair Cross-Domain Benchmark}%
}
\author{%
  \textbf{Ali Sultonov}\\
  Independent Researcher\\
  \texttt{troxtergrif@gmail.com}\\
  \url{https://github.com/Troxter222/psilogic}%
}
\date{}

\begin{document}
\maketitle

\begin{abstract}
Adaptive optimizers such as Adam and AdamW apply the same update rule regardless of
whether training is in a chaotic early phase or near convergence. We introduce
\textbf{PsiLogic} ($\Psi$Logic), an optimizer that augments Adam with a \emph{dynamic Active Cancellation
Term} gated by a dual exponential moving average (EMA) of scale-normalized gradient norms.
The resulting \emph{chaos detector} strengthens damping when gradient statistics are unstable
and fades to zero as training stabilizes, providing an implicit warmup without a hand-tuned
schedule.

We evaluate PsiLogic against Adam, AdamW, and Lion using \textbf{FairBench}---a reproducible
benchmark protocol with per-optimizer learning-rate sweeps, identical initialization per seed,
and Welch $t$-tests. On an NVIDIA H100 80GB reference run (4 arenas, 3 seeds, 2000 steps,
bf16 AMP), PsiLogic achieves the best validation metric in \textbf{three of four arenas}: NLP
perplexity $7.79 \pm 0.18$ vs.\ $8.17 \pm 0.08$ (AdamW, $p = 0.049$), ViT top-1 accuracy
$0.244 \pm 0.006$ vs.\ $0.223 \pm 0.002$ (AdamW, $p = 0.015$), and ResNet top-1 accuracy
$0.222 \pm 0.001$ vs.\ $0.172 \pm 0.004$ (Adam, $p = 0.001$). On diffusion, validation MSE
is statistically tied with Adam/AdamW ($p = 0.49$). ResNet accuracy vs.\ AdamW is a numerical
tie without significance at three seeds ($p = 0.44$). Peak GPU memory is comparable across
optimizers; PsiLogic incurs \textbf{1.2--1.8$\times$} wall-clock overhead on transformer-heavy arenas
(implementation-bound; Section~\ref{sec:limitations}).

We release an open-source PyTorch implementation, the full FairBench harness, and all raw
CSV outputs to support independent verification.

\end{abstract}

\section{Introduction}

The choice of optimizer affects convergence speed, generalization, and training stability in
deep learning. Adam~\cite{kingma2015adam} and AdamW~\cite{loshchilov2019adamw} dominate
practice, yet their corrective signal does not adapt to \emph{how confused the model currently is}.
At initialization, gradients are large and noisy; near convergence, they are small and stable.
Standard Adam treats both regimes with structurally similar updates.

We propose \textbf{PsiLogic}, which adds a chaos-conditioned damping term to the Adam update.
The term is strongest when a dual EMA of normalized gradient norms signals instability, and
vanishes automatically as training settles. PsiLogic is designed as a drop-in replacement
for \texttt{torch.optim.Adam} with optional task presets
(\texttt{PsiLogicNLP}, \texttt{PsiLogicGPT}, \texttt{PsiLogicViT}).

\paragraph{Contributions.}
\begin{enumerate}[leftmargin=*, itemsep=2pt]
  \item \textbf{PsiLogic optimizer}---chaos-gated Active Cancellation on top of Adam, with unified
    decay, optional gradient centralization (GC), and adaptive gradient clipping (AGC).
  \item \textbf{FairBench}---a bias-mitigated evaluation protocol: per-optimizer LR sweep, identical
    weights per seed, multi-arena tasks, and Welch $t$-tests.
  \item \textbf{Reference H100 benchmark}---reproducible CSVs and learning-curve plots committed at
    \texttt{benchmark/results/full/}, showing competitive or superior quality on NLP, ViT, and ResNet
    with explicit reporting of non-significant and negative results.
\end{enumerate}

We do \textbf{not} claim universal dominance over AdamW or Lion. We report limitations---including
step-time overhead and ties on diffusion and ResNet-vs-AdamW---explicitly.

\section{Related Work}
\label{sec:related}

\paragraph{Adaptive first-order methods.}
Adam~\cite{kingma2015adam} maintains bias-corrected first- and second-moment estimates for
per-parameter adaptive rates. AdamW~\cite{loshchilov2019adamw} decouples weight decay from the
gradient step and is the de facto standard for Transformers. AdaFactor~\cite{shazeer2018adafactor}
reduces memory via factored second-moment estimates. Lion~\cite{chen2023lion} uses sign-based
updates with coupled weight decay; it can be memory-efficient but often requires careful LR tuning.

\paragraph{Large-batch and layer-wise scaling.}
LARS~\cite{you2017lars} and LAMB~\cite{you2019lamb} rescale updates using the ratio of
parameter norm to gradient norm, stabilizing very large minibatch training. These methods
address scale mismatch across layers but do not gate damping on online gradient \emph{volatility}
the way PsiLogic's chaos detector does.

\paragraph{Second-order and curvature-aware methods.}
Shampoo~\cite{gupta2018shampoo} and Sophia~\cite{liu2023sophia} incorporate richer curvature
or Hessian information for faster convergence, at higher per-step cost. PsiLogic stays in the
first-order Adam family and adds only scalar chaos statistics shared across parameters.

\paragraph{Automatic learning-rate and warmup.}
Manual LR warmup~\cite{goyal2017large} is standard for large-batch SGD and Transformers.
Hypergradient descent~\cite{baydin2017hypergrad} differentiates through the optimizer to adapt
the LR online. Recent \emph{parameter-free} methods such as D-Adaptation~\cite{defazio2023dadapt}
and Prodigy~\cite{mishchenko2023prodigy} estimate a suitable global step size from observed
gradients. PsiLogic offers a complementary, chaos-driven \emph{implicit warmup}: effective
damping rises when gradient statistics are unstable and fades without an external schedule.

\paragraph{Stability mechanisms.}
Gradient centralization~\cite{yong2020gc} and adaptive gradient
clipping~\cite{brock2021nfn} improve training stability. PsiLogic optionally integrates both.
Its active-cancellation term is orthogonal: it shrinks weights when chaos is detected, rather
than only rescaling or clipping gradients.

\paragraph{Optimizer evaluation.}
Fair comparison requires matched tuning budgets. FairBench gives
each optimizer its own LR search rather than a single shared LR, reducing tuning bias that
has historically confounded optimizer comparisons.

\section{Method}

\subsection{Notation and Per-Group Hyperparameters}

PsiLogic operates on parameter groups indexed by $k$, each with learning rate $\eta$, AdamW weight decay
$\lambda$, chaos gain $\gamma$, and a group-specific \textbf{chaos amplification factor}
$P_k \ge 0$ (implementation name \texttt{p\_ext}, default $1.0$). $P_k$ lets presets assign
stronger cancellation to sensitive groups (e.g., embeddings) and weaker damping to others,
without changing the global chaos signal. We write $\mathbf{g}_t = \nabla_\theta \mathcal{L}$ for the
gradient at step $t$ and $w_t \in [0,1]$ for the \emph{chaos warmup weight}
(Section~\ref{sec:warmup}); this avoids overloading $\mathbf{g}_t$ with scalar gains.

\subsection{Update Rule}

Each step first applies unified multiplicative decay, then the bias-corrected Adam gradient step.
Composing the two operations yields
\begin{equation}
  \theta_{t+1}
  = \theta_t \cdot (1 - \delta_t)
    - \eta \cdot \frac{\hat{m}_t}{\sqrt{\hat{v}_t} + \varepsilon}\,,
  \label{eq:update}
\end{equation}
where $\hat{m}_t$ and $\hat{v}_t$ are the usual \emph{bias-corrected} Adam moments (we use $\hat{\cdot}$ consistently below and in Listing~\ref{lst:psilogic}) and $\delta_t$ is the
scalar unified-decay coefficient (Section~\ref{sec:decay}). Listing~\ref{lst:psilogic} spells out the
same sequence procedurally.

\subsection{Chaos Detector}

Let $\mathrm{gn}_t = \|\mathbf{g}_t\|_2 / \sqrt{\mathrm{numel}}$ be the scale-normalized gradient norm. We maintain:
\begin{align}
\mathrm{fast}_t &= 0.90 \cdot \mathrm{fast}_{t-1} + 0.10 \cdot \mathrm{gn}_t \\
\mathrm{slow}_t &= 0.99 \cdot \mathrm{slow}_{t-1} + 0.01 \cdot \mathrm{gn}_t \\
\mathrm{ratio}_t &= \mathrm{fast}_t / (\mathrm{slow}_t + \varepsilon) \\
\mathrm{chaos}_t &= \tanh(\mathrm{slow}_t) \cdot \bigl(1 + 0.5 \cdot \tanh(\relu(\mathrm{ratio}_t - 1))\bigr)
\end{align}
The fast and slow EMAs correspond to effective horizons of roughly $10$ and $100$ steps, respectively.
In adaptive mode (default), cancellation activates when $\mathrm{fast}_t > \tau_{\mathrm{scale}} \cdot \mathrm{slow}_t$
($\tau_{\mathrm{scale}} = 2.0$), detecting relative spikes in gradient chaos. As $\mathrm{slow}_t \to 0$ at
convergence, $\mathrm{chaos}_t \to 0$ and PsiLogic reduces toward AdamW-like behavior.

\subsection{Unified Decay}
\label{sec:decay}

Naively applying weight decay and active cancellation as separate multiplicative factors,
e.g.\ $\theta(1-\eta\lambda)$ followed by $\theta(1-c_t)$, would shrink parameters by
$(1-\eta\lambda)(1-c_t) \approx 1 - \eta\lambda - c_t$ only to first order; at large
early-step rates the cross term $-\eta\lambda c_t$ over-dampens weights. PsiLogic instead
computes one combined coefficient per step.

Define the chaos warmup weight $w_t \in [0,1]$ (ramps from 0 during an initial warmup window;
Section~\ref{sec:warmup}) and the per-group spike mask $s_t \in \{0,1\}$ from the chaos gate.
The raw cancellation fraction before clamping is
\begin{equation}
  \tilde{c}_t = s_t \cdot \mathrm{chaos}_t \cdot \eta \cdot \gamma \cdot P_k.
  \label{eq:raw-cancel}
\end{equation}
We clamp and add weight decay:
\begin{equation}
  \delta_t = \eta \lambda + w_t \cdot \min\!\bigl(\tilde{c}_t,\; c_{\max}\bigr),
  \qquad c_{\max} = \texttt{max\_cancel}
  \label{eq:delta}
\end{equation}
which is the $\delta_t$ used in Eq.~\eqref{eq:update}. Intuitively, $\eta\lambda$ is the AdamW decay contribution;
$w_t \cdot \min(\tilde{c}_t, c_{\max})$ is chaos-gated active cancellation scaled by $P_k$;
$c_{\max}$ (default 0.05) caps per-step shrinkage during volatile initialization.
Optional cosine schedules on $\gamma$ are supported via \texttt{gamma\_T\_max}.

\paragraph{Optional quantum decay.}
Let $q_0 \ge 0$ denote the \texttt{quantum\_decay} hyperparameter (default $q_0 = 0$ disables the feature).
When $q_0 > 0$, an effective rate $q_t$ is cosine-scheduled over training alongside $\gamma$
via \texttt{gamma\_T\_max} (the same schedule helper used for $\gamma$).
After computing $\delta_t$ but before the Adam subtraction in Listing~\ref{lst:psilogic}, each coordinate
is multiplied by
\begin{equation}
  \rho_{t,i} = 1 - \eta\, q_t\, w_t\, (1 - s_t)\, \tanh\!\bigl(|g_{t,i}|\bigr),
  \label{eq:quantum-rho}
\end{equation}
but only when $s_t = 0$, so auxiliary gradient-dependent regularization does not stack with
active cancellation on spike steps. FairBench presets use $q_0 = 0$ unless noted.

\subsection{Chaos Warmup}
\label{sec:warmup}

For \texttt{chaos\_warmup}$=-1$, the warmup horizon auto-scales as
$\max(500,\, T/20)$ over $T$ training steps. While $t \le t_{\mathrm{warm}}$, $w_t=0$; then
$w_t$ ramps linearly to $1$ over $t_{\mathrm{warm}}/4$ steps. This prevents the chaos term
from firing into raw from-scratch gradient noise.

\subsection{Algorithm}

\begin{lstlisting}[caption={PsiLogic (simplified; matches Eqs.~\ref{eq:delta}, \ref{eq:update}).},
                    label={lst:psilogic}]
for t = 1 ... T:
    grad <- nabla L(theta); optionally apply AGC and gradient centralization
    update Adam moments m, v; update fast_t, slow_t from ||grad||_2
    s_t <- spike mask from chaos gate
    c_t <- min(s_t * chaos_t * eta * gamma * P_k, max_cancel)
    delta <- eta*lambda + w_t*c_t
    theta <- theta * (1 - delta)          # unified decay
    # optional quantum decay when q_0 > 0:
    theta <- theta * (1 - eta*q_t*w_t*(1-s_t)*tanh(abs(grad)))
    theta <- theta - eta * m_hat / (sqrt(v_hat) + eps)   # Adam step
\end{lstlisting}
Full implementation: \url{https://github.com/Troxter222/psilogic} (\texttt{psilogic/optimizer.py}).

\subsection{Comparison with Baselines}

\begin{table}[t]
  \centering
  \caption{Feature comparison of optimizers.}
  \label{tab:features}
  \small
  \begin{tabular}{lcccc}
    \toprule
    Property & Adam & AdamW & Lion & \textbf{PsiLogic} \\
    \midrule
    Per-parameter adaptive rates & \cmark & \cmark & \xmark & \cmark \\
    Unified decay (with chaos) & \xmark & \xmark & \xmark & \cmark \\
    Chaos-aware damping & \xmark & \xmark & \xmark & \cmark \\
    Implicit early-phase damping & \xmark & \xmark & \xmark & \cmark \\
    Batched \texttt{foreach} CUDA kernels & partial & \cmark & \xmark & \cmark \\
    \bottomrule
  \end{tabular}
\end{table}

\section{FairBench Evaluation}

\subsection{Protocol and Arenas}

All headline numbers come from one reference run on \textbf{NVIDIA H100 80GB HBM3}
(PyTorch 2.4.1+cu124, CUDA 12.4), configuration frozen in
\texttt{benchmark/results/full/config.json}. FairBench uses a two-stage protocol:
(1)~per-optimizer LR sweep over 7 log-spaced rates from $10^{-5}$ to $10^{-2}$ (500 steps each);
(2)~evaluation at the best LR for 2000 steps with seeds $\{0,1,2\}$ and identical initialization.
Shared settings: batch 64, bf16 AMP, grad clip 1.0, cosine LR, 100-step warmup.
Four arenas cover NLP (Small GPT / TinyStories), ViT-Tiny on CIFAR-100,
ResNet-18 on Tiny ImageNet, and DDPM on CelebA $64^2$.
Full protocol and arena tables are in Appendix~\ref{app:protocol}.
PsiLogic uses fixed per-arena presets; \textbf{only LR is tuned}, as for all baselines.

\subsection{Main Results}

\begin{table}[t]
  \centering
  \caption{Main FairBench results (mean $\pm$ std over 3 seeds). Best per row in bold.}
  \label{tab:main}
  \small
  \setlength{\tabcolsep}{4pt}
  \begin{tabular}{llcccc}
    \toprule
    Arena & Metric & Adam & AdamW & Lion & \textbf{PsiLogic} \\
    \midrule
    NLP & Perplexity $\downarrow$ & $13.66 \pm 0.22$ & $8.17 \pm 0.08$ & $21.04 \pm 1.41$ & $\mathbf{7.79 \pm 0.18}$ \\
    NLP & Val loss $\downarrow$ & $2.614 \pm 0.016$ & $2.101 \pm 0.010$ & $3.045 \pm 0.068$ & $\mathbf{2.053 \pm 0.023}$ \\
    ViT & Val acc $\uparrow$ & $0.079 \pm 0.003$ & $0.223 \pm 0.002$ & $0.213 \pm 0.002$ & $\mathbf{0.244 \pm 0.006}$ \\
    ResNet & Val acc $\uparrow$ & $0.172 \pm 0.004$ & $0.219 \pm 0.005$ & $0.205 \pm 0.007$ & $\mathbf{0.222 \pm 0.001}$ \\
    Diffusion & Val MSE $\downarrow$ & $0.01987 \pm 0.00006$ & $0.01987 \pm 0.00006$ & $0.02175 \pm 0.00025$ & $0.02009 \pm 0.00045$ \\
    \bottomrule
  \end{tabular}
\end{table}

\noindent\textbf{Selected LRs.} NLP---all $3.16 \times 10^{-4}$; ViT---Adam $3.16 \times 10^{-5}$, AdamW/PsiLogic $3.16 \times 10^{-4}$,
Lion $10^{-4}$; ResNet---Adam/Lion $10^{-4}$, AdamW/PsiLogic $3.16 \times 10^{-4}$; Diffusion---
Adam/AdamW/PsiLogic $10^{-3}$, Lion $10^{-4}$.
Welch $t$-tests, compute costs, and per-seed breakdowns are reported in
Appendix~\ref{app:sig}--\ref{app:seeds}.

\subsection{Learning Curves and Overhead}

\begin{figure}[t]
  \centering
  \begin{subfigure}[b]{0.48\linewidth}
    \centering
    \includegraphics[width=\linewidth]{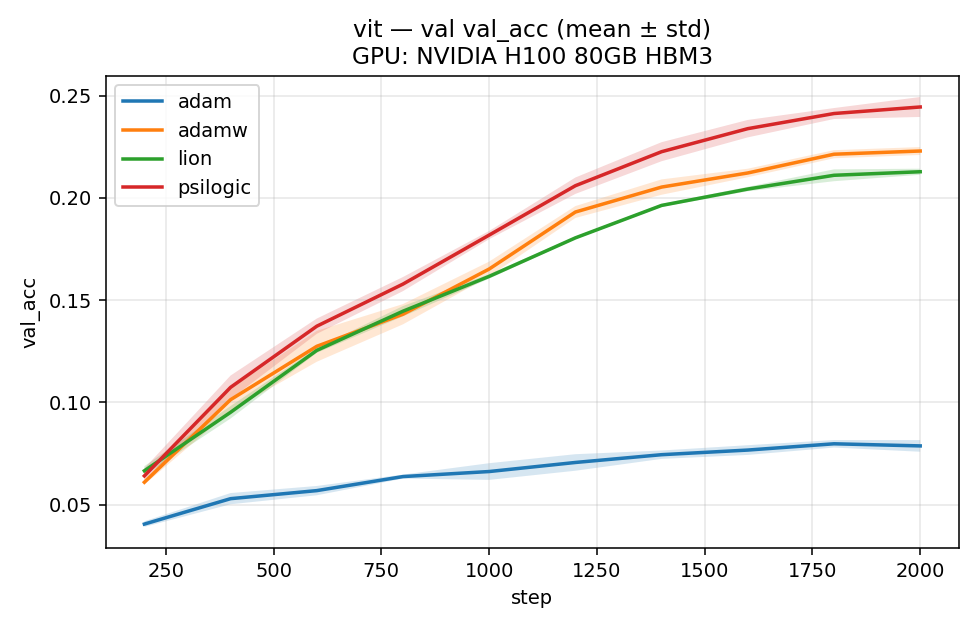}
    \caption{ViT val.\ accuracy.}
    \label{fig:vit}
  \end{subfigure}\hfill
  \begin{subfigure}[b]{0.48\linewidth}
    \centering
    \includegraphics[width=\linewidth]{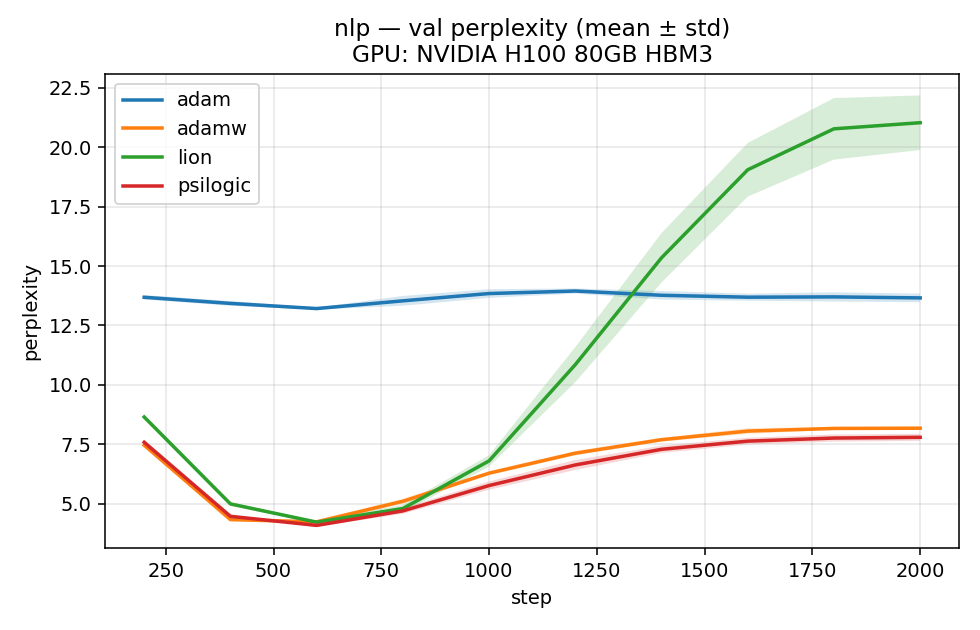}
    \caption{NLP perplexity.}
    \label{fig:nlp}
  \end{subfigure}\\[0.6em]
  \begin{subfigure}[b]{0.48\linewidth}
    \centering
    \includegraphics[width=\linewidth]{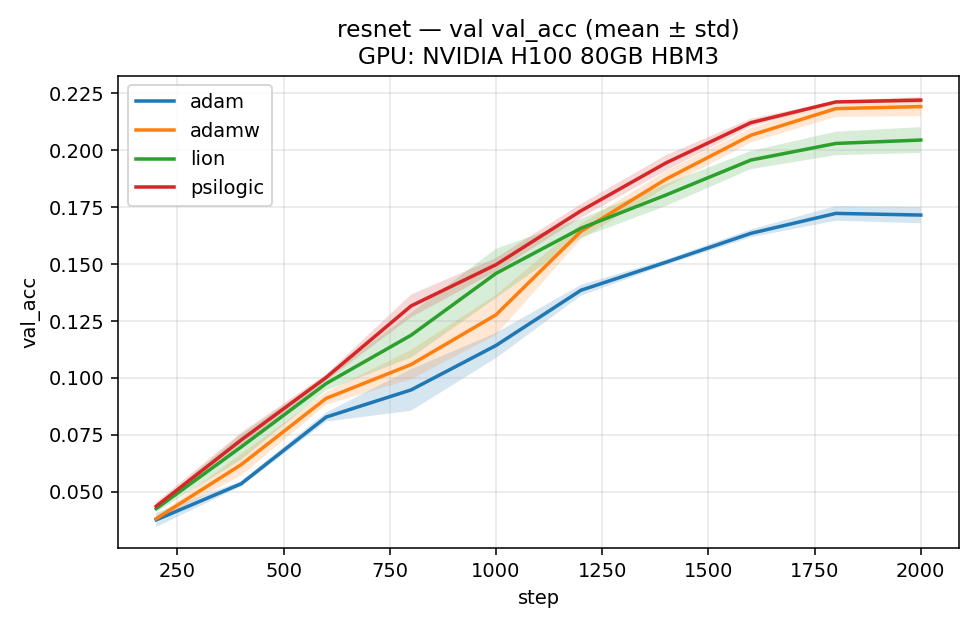}
    \caption{ResNet top-1 acc.}
    \label{fig:resnet}
  \end{subfigure}\hfill
  \begin{subfigure}[b]{0.48\linewidth}
    \centering
    \includegraphics[width=\linewidth]{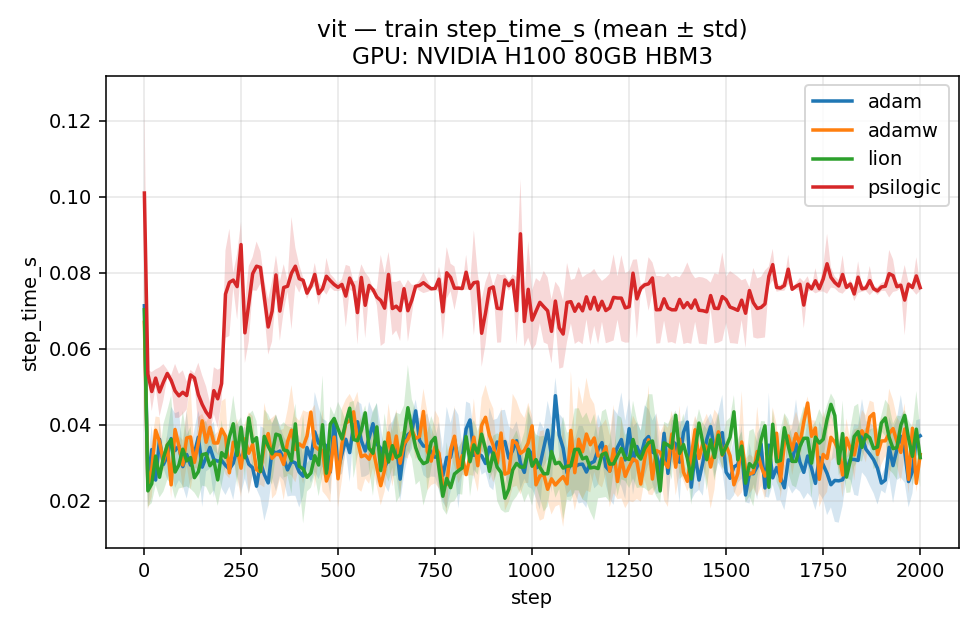}
    \caption{ViT step time.}
    \label{fig:overhead}
  \end{subfigure}
  \caption{FairBench learning curves (mean $\pm$ std) and ViT per-step wall-time overhead on H100.}
  \label{fig:curves}
\end{figure}

PsiLogic reports the best validation metric on NLP, ViT, and ResNet in Table~\ref{tab:main}.
Against Adam, all three gains are statistically significant (Appendix~\ref{app:sig}); against AdamW,
ViT and NLP perplexity are significant, while ResNet is a numerical edge ($0.222$ vs.\ $0.219$)
without significance at three seeds ($p=0.44$), and diffusion remains tied.
Step-time overhead reaches 1.79$\times$ on ViT (Figure~\ref{fig:curves}, panel~d); this gap is
implementation-bound rather than inherent to the chaos statistic (Section~\ref{sec:limitations}).

\section{Ablations and Component Analysis}

Prior ablations on a synthetic MLP task (v0.3.x) showed that gradient centralization and
adaptive gradient clipping each independently improve stability when combined with the chaos
term. A \emph{mirror ablation} demonstrated that dynamically mirroring PsiLogic's cancellation
magnitude as AdamW weight decay does not fully reproduce PsiLogic's per-parameter behavior,
indicating the chaos signal is not equivalent to a single global weight-decay schedule.

These ablations predate FairBench; component tests are maintained in \texttt{tests/}. Extended
FairBench ablations ($\gamma$, \texttt{max\_cancel}, \texttt{chaos\_warmup}) are planned.

\section{Discussion}

\paragraph{Why chaos damping helps.}
Large early gains on ViT ($0.244$ vs.\ $0.079$ Adam) suggest the chaos
term suppresses destructive early updates when gradient statistics are volatile. Under fair
LR tuning, NLP perplexity still favors PsiLogic over AdamW.

\paragraph{Implicit warmup.}
The cancellation term reduces effective step size during chaotic phases,
similar in spirit to LR warmup but driven by online gradient statistics rather than a fixed
schedule---complementary to hypergradient and parameter-free LR methods cited in
Section~\ref{sec:related}.

\paragraph{Reproducibility.}
ResNet shows the lowest cross-seed standard deviation among optimizers
($\pm 0.001$ on accuracy), which may matter for production training pipelines.

\paragraph{Limitations (stated explicitly).}
\label{sec:limitations}
\begin{enumerate}[leftmargin=*, itemsep=2pt]
  \item \textbf{Small seed count}---3 seeds; some comparisons (ResNet vs.\ AdamW, diffusion vs.\ AdamW) are not statistically significant.
  \item \textbf{Short training budget}---2000 steps per arena; not ImageNet- or LLM-scale.
  \item \textbf{Step-time overhead}---up to 1.79$\times$ vs.\ AdamW on ViT.
    The chaos detector tracks only scalar gradient statistics; the measured wall-clock gap is
    \emph{implementation-bound}---sequential element-wise PyTorch ops and the lack of fused
    \texttt{foreach} CUDA kernels in the reference build---not an inherent theoretical cost.
    Future releases can reduce overhead via kernel fusion (e.g., ATen or Triton).
  \item \textbf{Diffusion}---no quality win over Adam/AdamW at this budget.
  \item \textbf{No convergence proof}---empirical stability only.
  \item \textbf{Independent evaluation}---results have not yet been replicated by external groups.
\end{enumerate}

\section{Reproducibility Statement}

\begin{lstlisting}
git clone https://github.com/Troxter222/psilogic
cd psilogic && pip install -e ".[benchmark]" && pip install -r benchmark/requirements.txt
cd benchmark
python -m fairbench.download --data-root ./data
python -m fairbench --data-root ./data --output-dir results/full
\end{lstlisting}
Reference outputs: \texttt{benchmark/results/full/\{aggregate,summary,significance\}.csv}\\
Software DOI: \texttt{10.5281/zenodo.18739857} \quad PyPI: \texttt{pip install psilogic}

\section{Conclusion}

PsiLogic augments Adam with a chaos-gated Active Cancellation term that is strong during
unstable training and vanishes at convergence. Under FairBench on NVIDIA H100, it achieves
the best validation metric in three of four cross-domain arenas, with honest reporting of
ties and overhead. Future work: reduce step-time cost, increase seed count and training
length, and seek independent replication at scale.

\bibliographystyle{plain}

\appendix

\section{FairBench Protocol and Arenas}
\label{app:protocol}

\begin{table}[h]
  \centering
  \caption{FairBench protocol stages.}
  \label{tab:protocol}
  \small
  \begin{tabularx}{\linewidth}{lX}
    \toprule
    Stage & Description \\
    \midrule
    \textbf{Stage 1---LR sweep} & 7 log-spaced LRs from $10^{-5}$ to $10^{-2}$; 500 steps each; best val metric wins \\
    \textbf{Stage 2---Evaluation} & Selected LR; 2000 steps; seeds $\{0,1,2\}$; identical init per seed \\
    \textbf{Shared} & batch=64, bf16 AMP, grad\_clip=1.0, cosine LR, 100-step warmup \\
    \textbf{Statistics} & Mean $\pm$ std; Welch $t$-test (PsiLogic vs.\ each baseline) \\
    \bottomrule
  \end{tabularx}
\end{table}

\begin{table}[h]
  \centering
  \caption{FairBench arenas.}
  \label{tab:arenas}
  \small
  \begin{tabular}{llll}
    \toprule
    Arena & Model & Dataset & Metric \\
    \midrule
    NLP & Small GPT & TinyStories & Perplexity $\downarrow$, val loss $\downarrow$ \\
    ViT & ViT-Tiny patch16 224 & CIFAR-100 @ $224^2$ & Top-1 acc $\uparrow$ \\
    ResNet & ResNet-18 & Tiny ImageNet 200 & Top-1 acc $\uparrow$ \\
    Diffusion & DDPM + UNet & CelebA @ $64^2$ & Val MSE $\downarrow$ \\
    \bottomrule
  \end{tabular}
\end{table}

\section{Statistical Significance}
\label{app:sig}

\begin{table}[h]
  \centering
  \caption{Welch $t$-test: PsiLogic vs.\ baseline. $^{*}p<0.05$, $^{**}p<0.01$, $^{***}p<0.001$; n.s.\ = not significant.}
  \label{tab:sig}
  \small
  \begin{tabular}{llccc}
    \toprule
    Arena & Metric & vs Adam & vs AdamW & vs Lion \\
    \midrule
    NLP & Perplexity & $^{***}$ & $^{*}$ & $^{**}$ \\
    NLP & Val loss & $^{***}$ & n.s.\ ($p{=}0.054$) & $^{***}$ \\
    ViT & Val acc & $^{***}$ & $^{*}$ & $^{**}$ \\
    ResNet & Val acc & $^{**}$ & n.s.\ ($p{=}0.44$) & $^{*}$ \\
    Diffusion & Val MSE & n.s. & n.s. & $^{*}$ \\
    \bottomrule
  \end{tabular}
\end{table}

\section{Compute Cost}
\label{app:compute}

\begin{table}[h]
  \centering
  \caption{Compute cost on H100. A/W/L/P = Adam/AdamW/Lion/PsiLogic.}
  \label{tab:compute}
  \small
  \begin{tabular}{lccc}
    \toprule
    Arena & Peak VRAM (MB) A/W/L/P & Wall time (s) A/W/L/P & PsiLogic / AdamW \\
    \midrule
    NLP & 458 / 458 / 445 / 458 & 46.6 / 45.9 / 38.2 / 55.2 & 1.20$\times$ \\
    ViT & 1229 / 1229 / 1208 / 1229 & 95.2 / 98.5 / 98.6 / 176.7 & \textbf{1.79$\times$} \\
    ResNet & 823 / 825 / 777 / 823 & 45.3 / 47.6 / 46.1 / 67.4 & 1.42$\times$ \\
    Diffusion & 3780 / 3780 / 3768 / 3781 & 94.2 / 95.2 / 91.6 / 168.3 & \textbf{1.77$\times$} \\
    \bottomrule
  \end{tabular}
\end{table}

VRAM differences are $\le 3\%$ except Lion on ResNet/NLP (lower).

\section{Per-Seed Results}
\label{app:seeds}

\begin{table}[h]
  \centering
  \caption{Per-seed ViT validation accuracy.}
  \label{tab:vit-seeds}
  \begin{tabular}{rcccc}
    \toprule
    Seed & Adam & AdamW & Lion & \textbf{PsiLogic} \\
    \midrule
    0 & 0.078 & 0.226 & 0.214 & \textbf{0.238} \\
    1 & 0.083 & 0.222 & 0.211 & \textbf{0.247} \\
    2 & 0.076 & 0.221 & 0.213 & \textbf{0.249} \\
    \bottomrule
  \end{tabular}
\end{table}
Full per-seed tables for all arenas: \texttt{benchmark/results/full/summary.csv}.

\section{Archived Experiments}

Pre-FairBench results (CIFAR-10 A40, BERT, AG News, etc.) are archived in \texttt{OLD\_RESULTS.md}
and are \textbf{not} used for claims in this preprint.

\vfill
\arxivcomment{Ali Sultonov $\cdot$ Independent Researcher $\cdot$ arXiv preprint $\cdot$ \url{https://github.com/Troxter222/psilogic}}

\end{document}